# Forming Real-World Human-Robot Cooperation for Tasks with General Goal

Lingfeng Tao*, Michael Bowman *, Jiucai Zhang^, and Xiaoli Zhang*, *Member, IEEE*

*Abstract*— In Human-Robot Cooperation, the robot cooperates with humans to accomplish the task together. Existing approaches assume the human has a specific goal during the cooperation, and the robot infers and acts toward it. However, in real-world environments, a human usually only has a general goal (e.g., general direction or area in motion planning) at the beginning of the cooperation, which needs to be clarified to a specific goal (e.g., an exact position) during cooperation. The specification process is interactive and dynamic, which depends on the environment and the partners' behavior. The robot that does not consider the goal specification process may cause frustration to the human partner, elongate the time to come to an agreement, and compromise or fail team performance. This work presents the Evolutionary Value Learning approach to model the dynamics of the goal specification process with State-based Multivariate Bayesian Inference and goal specificity-related features. This model enables the robot to enhance the process of human's goal specification actively and find a cooperative policy in a Deep Reinforcement Learning manner. Our method outperforms existing methods with faster goal specification processes and better team performance in a dynamic ball balancing task with real human subjects.

## I. Introduction

Human-Robot Cooperation (HRC) is a promising topic in many applications, such as manufacturing [1], complex surgery [2] and autonomous driving [3]. For all HRC, the robot should adapt to the human to maintain healthy cooperation and help the human to achieve his/her goal [4]. Existing approaches in HRC assume the human has a specific goal at the beginning of the cooperation [5], where there is one or multiple targets (e.g., exact locations or objects) pre-known to the robot. The human may act toward one of these targets or change from one target to another. In these scenarios, the human always knows a specific target when he/she applies actions. However, this assumption may not be accurate in real-world applications. A common experience in real-life cooperation is that the team may hover around the expected area to find the best location/target/solution that satisfies all team members. We identified that *in realistic cooperation, the human may not have a specific goal at the beginning of the task but instead start with a general goal (e.g., a higher-level direction or area). The cooperation associates with a process that the goal is specified from general to specific (e.g., a target direction to a target location).* Furthermore, the goal specification and the corresponding cooperation formation are

*L. Tao, M. Bowman, and X. Zhang are with Colorado School of Mines, Intelligent Robotics and Systems Lab, 1500 Illinois St, Golden, CO 80401 USA (Phone: 303-384-2343; e-mail: tao@mines.edu, mibowman@mines.edu, xlzhang@mines.edu).

^J. Zhang is with the GAC R&D Center Silicon Valley, Sunnyvale, CA 94085 USA (e-mail: zhangjiucai@gmail.com)

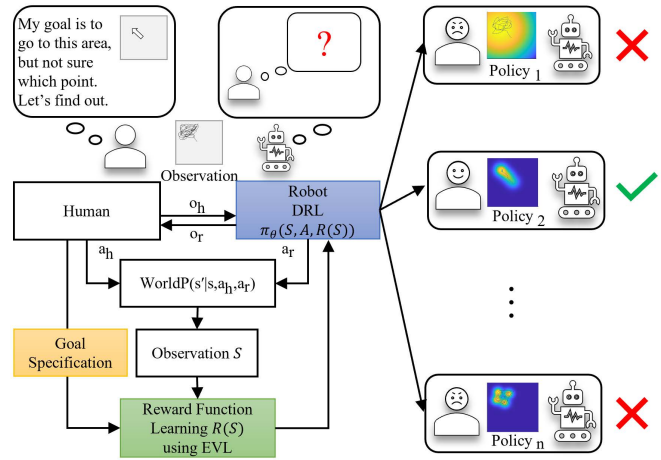

Figure 1. In realistic cooperation, the human may not have specific targets at the beginning of one task, but rather start with a higher-level goal (e.g., a general direction or area). In this work, the robot learns a cooperative policy in a Deep Reinforcement Learning (DRL) manner (blue block). The human specifies the goal with the help of the robot (yellow block). Then the observation and information of the goal specification process are used to learn a reward function using the proposed Evolutionary Value Learning algorithm. A good policy like Policy 2 helps the human to clarify the goal and achieve healthy cooperation. Bad policies like Policy 1 are too broad thus may frustrate the user, Policy n is too specific on data that it misses the human's actual goal.

interactive and dynamic, whose final equilibrium or agreement depends on the task and partner behaviors. Human actions are usually reactively chosen to specify the goal. Different cooperation processes can end up with different final equilibrium/agreements and different cooperation performances. *Thus, the robot needs to assist the human during the goal specification process to achieve better HRC when the human only has a general goal.*

In traditional HRC methods, the robot does not usually take the goal specification process into account because of the specific goal assumption. The robot is designed to identify one from the possible discretized goals, simplifying the goal inference to a classification problem, such as identify an object of interest to grasp or a specific way/routine for task execution. Then, the robot only needs to find which target is the human's goal and generate actions to assist the human. In contrast, a general goal is abstract and unclear. The human usually has no specific target for the robot to predict (classify), making it more difficult to form the cooperation.

The probability-based goal inference approaches, such as Bayesian inference [6], iteratively infer a human's goal based on observation. These studies still assume human has a specific goal; it is the robot who is not certain with the human's goal. So, it infers the human's goal with a form of probability fusion. However, when the human has no specific goal, the inference usually has low confidence and high variance because the human's reactive behaviors when specifying the goal may mislead the robot in the wrong direction. It also lacks the ability to memorize past human behaviors. If the human wants to recall an experienced potential target in the distant past, it is difficult for the inference model to follow.

The learning-based approaches like Inverse Reinforcement Learning (IRL) [7] learn the human goal from the demonstration (e.g., trajectory). However, with a general goal, the human may end up with different task/cooperation equilibriums in different trials in the goal specification process, which are inconsistent for learning. In such conditions, current HRC methods are limited when modeling or inferring the human's general goal, resulting in the human's adaption or correction to the robot's behavior rather than vice versa.

This work presents an Evolutionary Value Learning (EVL) approach to first build an abstract goal inference model from the observation with a State-based Multivariate Bayesian Inference (SMBI) method. SMBI model continuously covers the environment's state space, which does not require discretized representation of specific targets. Meanwhile, the robot experience is then used to extract the goal specificity-related features describing the human's historical goal specification process. Finally, EVL uses the probability distributions of these features to iteratively shape the SMBI model to build an evolutionary value model of the human's general goal. The global perspective of the evolutionary value model can naturally turn to a reward function for the robot to solve the HRC under a reinforcement learning manner. The contribution of this work is summarized as follow:

1) **Development of an Evolutionary Value Learning (EVL) approach** that includes

a) **A State-based Multivariate Bayesian Inference (SMBI)** to online model the dynamics of the goal specification process with a global perspective.

b) **An Evolutionary Value Updating** method to actively generate the evolutionary reward function with the SMBI model and offline feature extraction to enhance the process of goal specification and cooperation formation.

2) **Validation of the EVL approach** with human subjects in a dynamic task with a straightforward visualization of the human-robot cooperation and goal specification process.

## II. RELATED WORK

Conventional human goal prediction methods build inference models on three types of objectives. The first type is the target-based objective, where the robot infers the human's target and helps the human reach the target, such as an object grasping task [8]. The second type is routing-based objective, where the robot infers what routing the human is following and helps to finish the rest of the steps, such as a cooperative cooking task [9]. The third one is the action-based objective, where the robot predicts which action the human will execute, such as in a table-carrying task [10]. The robot needs to predict the human's rotational and translational direction. Overall, the above approaches are built under the assumption that the human has a specific goal when deciding the object of interest, executing the routing, or generating an action plan, which may not work appropriately if the human only has a general goal.

Technically, literature has used Bayesian Inference (BI) methods to build target-based human goal predictive models from human behavior observation. BI iteratively updates the posterior probability of the human's goal according to the observed information. For example, a multi-class BI model can be used to predict the human's goal of grasping [11]. BI was also used to learn a transition function to describe how humans select a target [12]. This approach infers the goal with partially observed human behavior, which still assumed that the human has a specific goal among several known targets. A potential problem of current Bayesian approaches is that the posterior is updated with the online observations. However, when the human goal is not specified, real-time human behavior and state may not sufficiently convey the goal's information during the earlier specification process. Furthermore, when the human wants to recall a visited target in the distant past, the updating mechanism of BI methods cannot quickly follow the human due to its inability to memorize sparse information [13].

Learning-based methods are popular in the last decade to enable robots to understand the human goal. Literature has used IRL in HRC to build empirical reward functions based on human demonstrations [14]. The empirical reward function is then used in a Reinforcement Learning (RL) setting to guide the policy learning process. Recent approaches such as Cooperative IRL [15] (also known as value alignment), shows that human can actively teach the robot to cooperate in HRC. However, such learn-from-demonstration approaches require humans to act optimally with dependable demonstrations (i.e., a specific goal) [16]. Although some methods like [17] allow the human to have different ways to achieve the goal (e.g., different routes), it is still assumed that the goal is specific because the final states are the same. Thus, the IRL method is unlikely feasible when the human has a general goal.

Human-in-the-loop algorithms attempt to learn human value while robots and humans interact with the environment [18], such as a cooperative assembly task [19]. Literature has used a reinforcement function [20] to model the relationship between the state-action pairs and the human's values and further to be learned by the RL agent. Human-in-the-loop algorithms assume the human can give optimal actions, and the human's role is to guide the robot to learn the policy. This work focuses on enabling the robot to help the human specify the goal together rather than the other way around.

## III. EVOLUTIONARY VALUE LEARNING

### A. Model Structure

We model the HRC as an RL problem that follows the Markov Decision Process (MDP) [21]. The MDP is defined as a tuple $\{S, A, R, \gamma\}$, where $\{a_h, a_r\} \in A$ is the set of actions of the human and the robot, $R(s_{t+1}|s_t, a_t)$ is the reward function that gives the reward after a transition from state $s_t$ to state $s_{t+1}$. $\gamma$ is a discount factor. A policy $\pi(s, \theta)$ specifies an action for state $s$ with the network parameters of $\theta$. PPO algorithm [22] is adopted to find $\theta$. The following sections introduce SMBI and EVL to construct and update the reward function based on observation (section B) and information of the human goal specification process (section C).

### B. State-based Multivariate Bayesian Inference (*SMBI*)

When a human has a general goal, prior knowledge is not available for the robot to generate an initial policy. It must learn to cooperate during the task. Thus, a method that can model humans in real-time is needed. Multivariate Bayesian

Inference (MBI) has been recently used to model the dynamic of human mental behaviors in robotics. In [23], MBI was first used to model human preference with online observation. Then it was adopted in [24] to optimize an exoskeleton based on the user feedback. This section introduces SMBI, which builds a multivariate inference model of the human goal based on the observed human's exploration in the state space. We denote an observation trajectory as $T=\{s_1,...,s_t,...,s_\Gamma\}$, where $\Gamma$ is the length of the trajectory. $s_t=[d_1,...,d_i,...,d_n]_t$ is the state vector at $t$ time step, where $d_i$ is the state component, $n$ is the length of the state vector. We assume there exists a latent, underlying goal $g(d_i)$ for the human (but not yet specified) for each state component $d_i$. So we write the unified goal as a vector $F=[g(d_1),...,g(d_i),...,g(d_n)]^{Tr}$, where $Tr$ is the transpose operator. Given the state $s_t$, we want to find the posterior probability of $F$:

$$P(F|s_t) \propto P(s_t|F)P(F) \quad (1)$$

we define the multivariate Gaussian prior over $F$:

$$P(F) = \frac{1}{\sqrt{(2\pi)^n|\Sigma|}} exp\left(-\frac{1}{2}(F)^{Tr}\Sigma^{-1}(F)\right) \quad (2)$$

where $\Sigma \in R^{n \times n}$ is the covariance matrix, the $xy$-th element of $\Sigma$ is a Gaussian kernel:

$$K(d_x, d_y) = exp\left(-\frac{\Gamma}{2}\sum_{c=1}^{n}(d_x^c - d_y^c)^2\right) \quad (3)$$

The likelihood is the joint probability of observing the state $s_t$ given the latent function values, which is calculated as:

$$P(s_t|F) = \prod_{i=1}^{n} P(d_i|g(d_i))|t \quad (4)$$

From the Bayes' theorem, the posterior $P(F|s_t)$ is:

$$P(F|s_t) = \frac{P(F)}{P(s_t)} \prod_{i=1}^{n} P(d_i|g(d_i))|t \quad (5)$$

Where the prior $P(F)$ is defined in Eq. 2, the likelihood function is defined in Eq. 4, the normalization factor $P(s_t) = \int P(s_t|F)P(F)dg$. The posterior $P(F|s_t)$ can be estimated with the Laplace approximation method [25]. The learned inference model fully covers the environment's state space. Thus, the learned probability distribution of goal is suitable to act as the reward function for the RL training [26].

*C. Evolutionary Value Learning*

Pure Bayesian methods is highly depended on the online sample, which may not be ideal for a stable learning process, as explained in section 1. This section introduces the EVL method, which builds an evolutionary reward function $R$ with the posterior of the human goal $P(F|s_t)$ from the SMBI model and the probabilistic human guidance features $\{H_1,...,H_m,...,H_M\}$, where $M$ is the number of features. Each feature is a multivariate distribution over the state component $d_i$, which has the same dimensions as the SMBI model. In this work, we define the following features:

1) Spectral entropy [27], which can measure human behaviors' spectral power distribution or information density. The distribution of the spectral entropy is defined as:

$$H_1(d_i)_t = -\sum_{i=1}^{\Gamma} d_i log_2 P(d_i)|t \quad (6)$$

where $P(d_i) = \frac{|X(d_i)|^2}{\sum |X(d_i)|^2}$ is the probability distribution, $X(d_i)$ is the discrete Fourier transform of $d_i$. This feature aims to encourage the robot to explore the high entropy (i.e., high information) state based on human exploration.

2) The visiting frequency of state gives the first-order information on how the human explores [28]. For example, the human may revisit the states that may give the best reward. The robot should go to the states that are explored most of the time. Then, the probability distribution of the state component $d_i$ is defined to follow a kernel probability density function:

$$H_2(d_i) = \frac{1}{\Gamma}\sum_{t=1}^{\Gamma} K(d_i)|t \quad (7)$$

where $K$ is a uniform kernel function. $t$ is the time step.

3) The human exploration pace gives the second-order information on how fast the human explores the environment. The human may search slower in high potential states and faster in low potential states. The robot should follow the human's pace. This feature is designed as the probability distribution of the derivative of the state component $d_i$:

$$H_3\left(\frac{d_i - d_{i-1}}{\tau}\right) = \frac{1}{\Gamma}\sum_{t=2}^{\Gamma} K\left(\frac{d_i - d_{i-1}}{\tau}\right)|t \quad (8)$$

where $\tau$ is the sample time.

4) The human reaction reveals the third-order information on how the human reacts in the environment. For example, when the object moves to a dangerous zone or the robot is executing counterintuitive action, humans may show a fast and strong reaction that causes high acceleration. If the object is safe and the team cooperates well, the human reaction is soft such that the object's acceleration is slow. Thus, the robot also needs to behave like a human. This feature is defined as:

$$H_4\left(\frac{d_i - 2d_{i-1} + d_{i-2}}{\tau^2}\right) = \frac{1}{\Gamma}\sum_{i=3}^{\Gamma} K\left(\frac{d_i - 2d_{i-1} + d_{i-2}}{\tau^2}\right) \quad (9)$$

Then we construct the reward function with Eq. 5 – Eq. 9:

$$V = \alpha P(F|s_t) - \eta + \sum_{m=1}^{M} \beta_m H_m(s_t) \quad (10)$$

where $H_m(s_t) = \prod_{i=1}^{n} H_m(d_i)$, which is the joint probability distribution across the state vector. $\alpha \in R$ is a factor in changing the shape of the posterior. $\beta_m$ are the linear blending weights to scale the features to the same decimal as the posterior and fine-tune each feature's contribution. In this work, the designed weights $\beta_m$ for each featurs are shown in Table 1. $\eta \in R$ is a parameter to control the magnitude of the positive reward region. The robot collects information about the goal specification process during the cooperation (state trajectories). As a result, the reward function is evolving to describe better the human's value of the goal that is being specified by both the human and the robot. Thus, we call the updating process EVL, and its updating process is shown in algorithm 1.

---

**Algorithm 1** EVL

1: **procedure** Initialize agent policy $\pi_0 = \pi(\theta_0)$, initialize goal prior $P(F)$
2:   **for** *episode=1,2,...* **do**
3:     obtain initial state $s_1$
4:     **for** *t=1,2,...Γ* **do**
5:       Select action $a_t$ with policy $\pi(s_t, \theta_{t-1})$
6:       **return** trajectory $[s_1, a_1, s_2, a_2, ..., s_\Gamma, a_\Gamma]$
7:   **end for**
5:   Update posterior $P(F|s_t)$ with Eq.1 – Eq. 5
8:   Update $H_m \; \forall m=1,2,3,4$ with Eq.6 – Eq. 9
9:   Update evolutionary reward function $V_t$ with Eq. 10.
10:   Update policy $\pi(s_t, \theta_t)$
11:   **if** human stop = true **break**
12: **end procedure**

## IV. EXPERIMENTS

### A. Setup

A ball rolling task was designed (Fig. 2), consisting of a ball on the board. The human and the robot can rotate the board to make the ball to roll. The setup accepts multiple players to operate. The ball's movement is the visualizable mark of the human-robot cooperation formation and goal specification process. The experiment covers dynamic target behavior and indirect control of the target by adjusting the board pose and the indirect interaction between players, which effectively evaluates the EVL with its various dynamics and uncertainties.

Two environments were designed. Environment 1 (Fig. 3a) evaluates the effectiveness of EVL to learn the dynamics of the human and the robot's cooperation. It contains a rectangular board with four walls and shifted pivots for both robots. Environment 2 (Fig. 3b) evaluates how the environment's dynamics affect human and robot cooperation. It uses a square board with two sides of the walls removed. The human and the robot need to find a safe position to prevent the ball from hitting the wall and falling off the board. In both environments, the human only knows the rough area she/he feels comfortable in but does not know the exact safe position because of the unknow dynamics of the environment. 15 human subjects are invited with the following rules: 1) the human subjects have no prior experience, and 2) the human subjects are not informed with any initial goal position nor the policy of the robot partner. The human subject can stop the cooperation if he/she thinks the ball stays in a safe position. The experiments were approved by the Institutional Review Board (IRB) of Colorado School of Mines. Prior to participating in the study, a short introduction was provided to the participants, including technologies involved, the system setup, and the purpose of the study.

For the RL agent, we define the state as $s=(x,y)$, which is the 2-D coordinates of the ball related to the board. Based on (4), the human goal's prior probability is following a 2-D Gaussian distribution. The experience is the trajectory of the ball. For both the human and the robot, the actions are roll ($x$ axis) and pitch ($y$ axis) of the board. We developed and compared the performance of two agents that are trained with the SMBI model, one (denoted as EVL) with and the other (denoted as Bayes) without consideration of the human guidance feature, respectively. A baseline model is developed with a pretrained policy that keeps the ball in the initial position, which is the midpoint of the two pivots. The policy doesn't update during the training (denoted as Fixed).

The environment is first implemented in CoppeliaSim [29], a virtual robot platform with integrated physical simulation. Two robots are used. One of the robots is controlled by the human, and the RL agent controls the other. After the simulation validation, EVL was tested in a real-world environment with one Kinova MICO arm. One human subject directly holds the other side of the board. The ball position is captured with a webcam at 30Hz using a pattern matching method [30] to achieve real-time processing.

### B. Evaluation Metrics

To evaluate EVL in the goal specification process and team performance, we define the following metrics. The first metric $(\mu_x,\mu_y)$ and second metric ($U$) are defined to evaluate the goal specification process, and the rest are defined for the team performance, which includes task performance ($L$ and $\delta$) and cooperate performance ($\sigma, \varphi$) after the training.

1) The ball trajectory's mean position for each training iteration gives an implicit prediction of the goal position. The mean position is calculated as

$$(\mu_x,\mu_y) = \left(\frac{1}{\Gamma}\sum_{t=1}^{\Gamma} x_t, \frac{1}{\Gamma}\sum_{t=1}^{\Gamma} y_t\right) \quad (11)$$

2) The goal specificity for each training iteration, which is the cumulative divergence (or variance) between the trajectory and the goal position (smaller is better). Its shows the goal changes from general to specific in training.

$$U = \sum_{t=1}^{\Gamma} \sqrt{(x_t-\mu_x)^2+(y_t-\mu_x)^2} \quad (12)$$

3) The total ball trajectory length (smaller is better).

$$L = \sum_{t=2}^{\Gamma} \left(\sqrt{(x_t-x_{t-1})^2+(y_t-y_{t-1})^2}\right) \quad (13)$$

4) The ball's density ratio is within 5% range around the goal position (larger is better).

$$\delta = \frac{\sum f_r(x,y)}{\Gamma}, \quad 0 \leq \delta < 1 \quad (14)$$

where $f_r(x,y)=\begin{cases}1, & \text{if } \rho<0.05\rho_{max} \\ 0, & f_r(x,y)=0.\end{cases}$, $\rho=\sqrt{(x-\mu_x)^2+(y-\mu_y)^2}$ is the distance between the position (x,y) and the mean $(\mu_x,\mu_y)$.

5) The cumulative effort (magnitude of actions) that the human executed (smaller is better).

$$\sigma = \sum |a_h| \quad (15)$$

where $a_h$ is the human's action command (i.e., rotation angle), which can be directly accessed and quantified in the simulation environment from the controller's analog input.

6) The ratio of agreement in the actions (larger is better).

$$\varphi = \frac{\sum ag}{\text{number of actions}}, \quad 0 \leq \varphi \leq 1 \quad (16)$$

where $ag = \begin{cases}1 & a_h*a_r>0 \\ 0 & a_h*a_r \leq 0\end{cases}$. $ag$ means agreement, which is 1 when human action and robot action are in the same direction. Otherwise, $ag$ is 0.

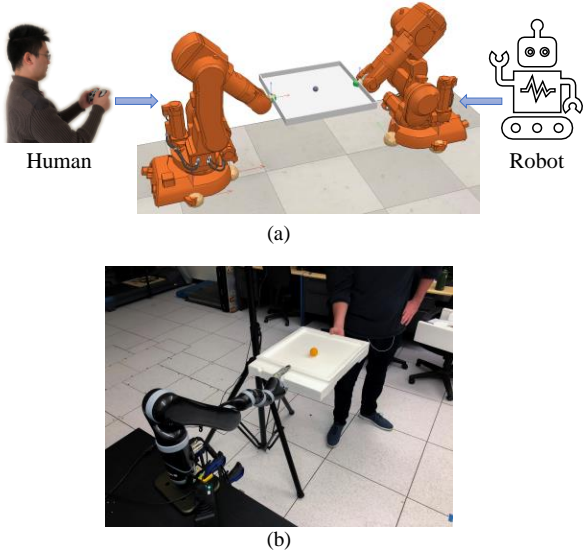

Figure 2. A ball rolling task in simulation (a) and real-world (b).

## V. RESULTS

### A. Enhanced Goal Specification Process with EVL

The hyper-parameters of the EVL and PPO agent are shown in Table I. For the PPO agent, the discount factor, clip factor, GAE factor, learning rate, and the epoch number are the same in both environments, which follow the default setup of the PPO algorithm. The sample time in the physical environment is twice as the simulation to take into count of the action exection time of the physical robot. The episode steps for simulation and physical experiment are correspondingly adjusted to set the episode length to 40 seconds, which is empirically determined in the tuning process. A longer episode length may cause distractions to the human subject. A shorter episode length may cause low learning efficiency. The training results from simulation and physical experiments are consistent in terms of the performance rank for the three methods. Compared to simulation, physical experiments took 3-5 times to converge due to the complexity and uncertainty of the physical environment. Fig. 3 shows the visualization of the training process for both environments in simulation. Fig. 4 shows the visualization of the training process for both environments in physical experiments.

The data from the simulation are used for quantitative analysis for its cleaner quality. The traces of the mean position $(\mu_x, \mu_y)$ during the goal specification processes for each training iteration are shown in Fig. 3c-4d. In environment 1, Fig. 3a shows the human wants to move the ball away from the initial position along the Y-axis. The robot only needs to update its policy related to the Y-axis and keep the X-axis's partial policy. In environment 2, Fig. 3b shows the human tries to find a safe position in the lower-left corner of the board to avoid the ball falling off and hitting the wall. The robot needs to update the policy in both the X-axis and the Y-axis.

TABLE I. HYPER-PARAMETERS OF EVL AND PPO AGENT

| Parameters | Simulation | Physical Experiment |
|---|---|---|
| Discount Factor($\gamma$) | 0.995 | 0.995 |
| Experience Horizon | 512 | 256 |
| Entropy Loss Weight | 0.02 | 0.04 |
| Clip Factor | 0.05 | 0.05 |
| GAE Factor | 0.95 | 0.95 |
| Sample Time | 0.05 | 0.1 |
| Episode Step | 800 | 400 |
| Mini-Batch Size | 64 | 128 |
| Learning Rate | 0.001 | 0.001 |
| Number of Epoch | 3 | 3 |
| Weight $\alpha$ | 10000 | |
| Weight $\beta_1$ for $H_1$ | 1000 | |
| Weight $\beta_2$ for $H_2$ | 10000 | |
| Weight $\beta_3$ for $H_3$ | 10000 | |
| Weight $\beta_4$ for $H_4$ | 10000 | |
| $\eta$ | 10 | |

The goal specificity (cumulative divergence) at each training iteration is shown in Fig. 3e-3f. For environment 1, the specificity (U=141) reached by EVL outperforms the Bayesian method (U=165) by 24% and Fixed policy (U=193) by 37%. For environment 2, the specificity (U=127) reached by EVL outperforms the Bayesian method (U=147) by 16% and baseline Fixed policy (U=207) by 63%. In both environments, the faster goal specification processes of EVL confirm that when humans only have a general goal, the robot needs to utilize the information in the historical observation to assess human behaviors comprehensively. The Bayesian inference method performs worse but still could manage to narrow down the goal specificity in both environments. While the human subject was still familiar with the environment and the partner, the Bayesian method mistook the latest observation as the human goal. When the human subject tried to move to the specified goal, it cannot follow and causes a performance drop. Thus, it has more performance oscillation throughout the learning process. The RL method with a fixed policy performs worst because no matter how the human acts, the robot always wants to follow its policy.

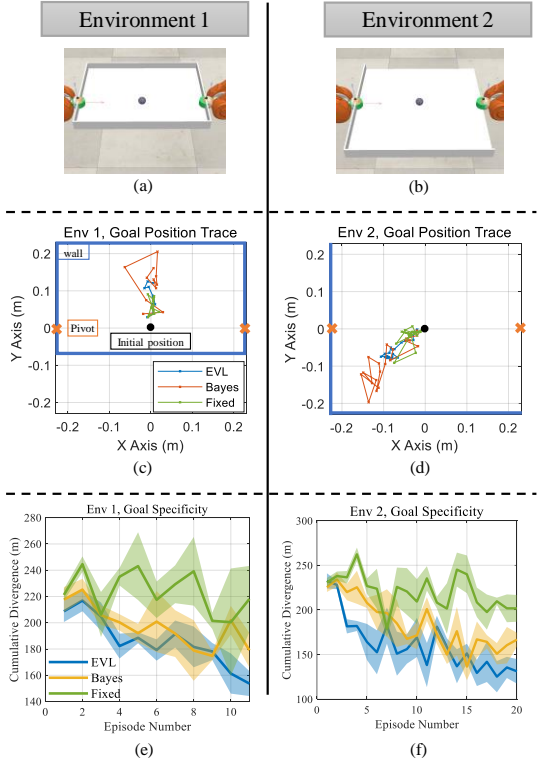

Figure 3. Enhanced goal specification processes with EVL in two environments (left column for environment 1, right for environment 2). (a)-(b) Environment setups. (c)-(d) The goal position trace of the ball trajectory for each iteration of the learning process. (e)-(f) The change in mean and variance of the goal specificity during the goal specification process for all subjects.

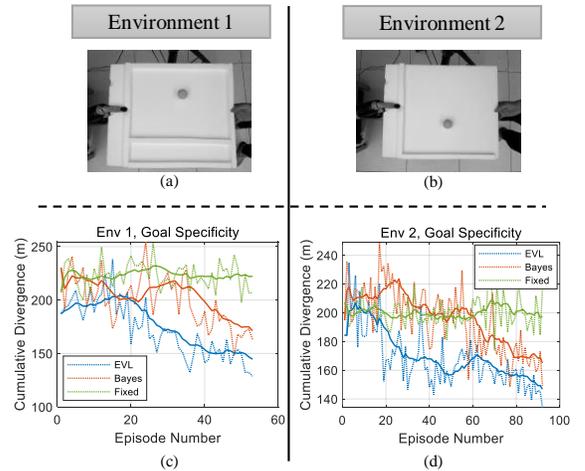

Figure 4. Enhanced goal specification processes with EVL in two physical environments (left column for environment 1, right for environment 2). (a)-(b) Environment setups. (c)-(d) The change in mean and variance of the goal specificity during the goal specification process.

Fig. 5 shows an example of the reward function updating process for environment 2. The robot has an initial policy. The cooperation starts at the time $t_0$. As the cooperation continues, at each sample time $t_i$, the robot utilizes the SMBI method and the goal specificity-related feature masks to generate new reward functions and update its policy. The linear blending weights for each feature mask are tuned to build an appropriate reward function. The final reward function represents the specified human goal is learned at time $t_\Gamma$. The features are designed from literature, which can be expandable and flexible to extract helpful information from the states and states' derivatives. Fig. 5 gives an insight into why EVL performs better than the baseline methods. The posterior $P(F|s_t)$ that learned from the SMBI method has a distinctive contour compared to the prior $P(F)$. If the agent learns from the updated posterior, the agent cannot update the policy quickly enough to follow the reward function. The robot may spend more time searching for a valid gradient to the goal position. On the contrary, the evolutionary reward function built with EVL connects the contour of the before the posterior. It provides a smooth and continuous value gradient for the agent to move to the goal position quickly.

### B. Enhanced Team and Cooperation Performance

Fig. 6 shows the validation of the team performance after the goal specification process. The human and robot cooperate for 40 seconds. During this cooperation process, the robot policy does not update. Fig. 6a-6d are the results for environment 1, including the ball trajectories for the three methods and the corresponding visualized reward functions. Fig. 6e-6h are the results for environment 2. The three methods' different goal positions show that different cooperation processes lead to different goals and task performance. Specifically, the goal positions of EVL and Bayesian inference are away from the initial position, which indicates that the robot has adapted to the human during the cooperation. The fixed policy's goal positions are closer to the initial position, in which the human had to accommodate more to the robot because the robot cannot adapt. In both environments, the ball trajectories of EVL are tightly distributed around the goal position. The ball trajectories of the baseline methods are more sparsely distributed around the goal position. The reciprocating behavior means there exist more disagreements caused by insufficient assistance.

Table II shows performance statistics to compare team performance under three cooperation methods. Overall, EVL performed better in environment 2 than environemnt 1 but took a longer time to specify the goal. On average of two environments, the cooperation achieved with EVL has the shortest trajectory (L=2.79m) and the highest density ratio (δ=0.305). The human working with the EVL agent spent the least effort (σ=41.71). EVL also achieved the highest agreement ratio (φ=0.66). The longer trajectories and higher human efforts of the Bayesian inference method are because the human struggles with the robot's insufficient assistance and spends more effort to correct them; it still helped the human to specify the goal and reduce the disagreement ratio and the number of human yields compared with the fixed policy method. Due to the fixed policy, the human had to adapt to the robot or even release control. It is the reason why it had shorter trajectories and lower human effort than the Bayesian inference method. However, its disagreement in action and the number of human yields are the highest.

TABLE II. PERFORMANCE STATISTCIS

| Env | Methods | Task Performance | | Cooperate Performance | |
|---|---|---|---|---|---|
| | | $L$ (m) | $\delta$ | $\sigma$ | $\varphi$ |
| 1 | EVL | 3.44±0.34 | 0.23±0.06 | 48.66±3.21 | 0.68±0.12 |
| | Bayes | 4.84±0.42 | 0.16±0.09 | 57.59±4.56 | 0.63±0.19 |
| | Fixed | 3.82±0.67 | 0.18±0.07 | 53.79±6.28 | 0.45±0.13 |
| 2 | EVL | 2.14±0.26 | 0.40±0.07 | 34.75±4.69 | 0.64±0.17 |
| | Bayes | 5.15±0.37 | 0.14±0.05 | 50.82±7.15 | 0.59±0.22 |
| | Fixed | 4.78±0.72 | 0.11±0.06 | 47.04±6.84 | 0.40±0.16 |

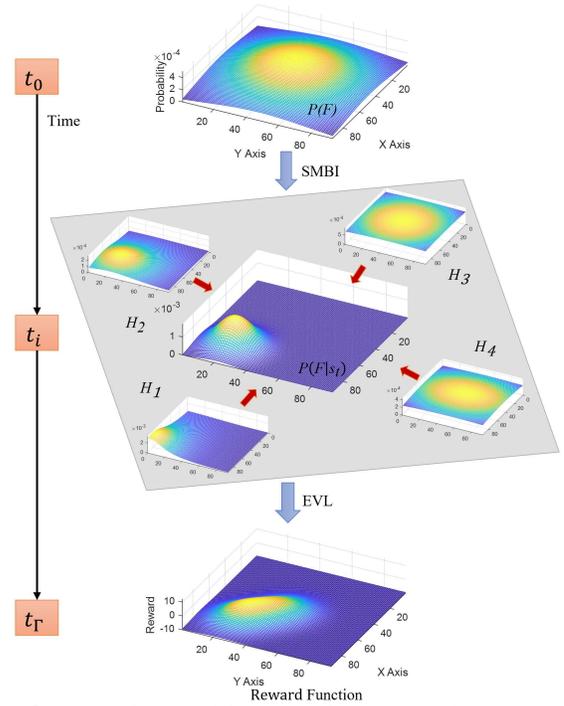

Figure 5. Visualization of the EVL updating process for environment 2, which starts from $t_0$ and ends at $t_\Gamma$. The figures in the middle are snapshots of the probability distribution of goal specificity related feature masks at $t_i$.

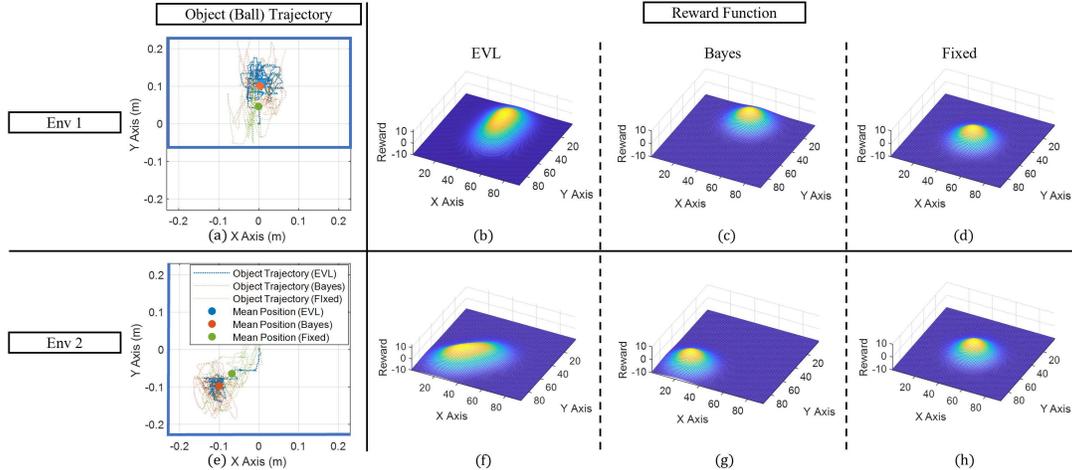

Figure 6. The object (ball) trajectories and visualizations of the reward functions after the goal clarification process. The top row is for environment 1, and the bottom row is for environment 2.

## VI. Conclusions

This work identified an unaddressed HRC problem where humans only have a general goal. It developed the EVL approach that integrates a novel SMBI algorithm for human general goal modeling and an EVL algorithm that extracts and utilizes the human guidance features from the historical observations for enhanced goal specification and cooperation formation. EVL was evaluated in both simulation and physical environment with real human subjects. The results prove EVL successfully accelerates the cooperation formation and smoothly helps humans to specify the goal. Our future work will focus on broadening the EVL to more complex tasks with partially observable environments.